# PACEMAKER: INTERMEDIATE TEACHER KNOWLEDGE DISTILLATION FOR ON-THE-FLY CONVOLUTIONAL NEURAL NETWORK

*Wonchul Son[1], Youngbin Kim[1], Wonseok Song[2], Youngsu Moon[2,] Wonjun Hwang[1]*

[1]Department of Computer Engineering, Ajou University, Republic of Korea
[2]Department of Visual Display, Samsung Electronics CO., LTD, Republic of Korea

**ABSTRACT**

There is a need for an on-the-fly computational process with very low performance system such as system-on-chip (SoC) [1] and embedded device etc. This paper presents pacemaker knowledge distillation (PMKD) as intermediate ensemble teacher to use convolutional neural network (CNN) in these systems. For on-the-fly system, we consider student model using 1×N shape on-the-fly filter and teacher model using normal N×N shape filter. We note three points about training student model, caused by applying on-the-fly filter. First, same depth but unavoidable thin model compression. Second, the large capacity gap and parameter size gap due to only the horizontal field must be selected not the vertical receptive. Third, the performance instability and degradation of direct distilling. To solve these problems, we propose intermediate teacher, named pacemaker, for an on-the-fly student. So, student can be trained from pacemaker and original teacher step by step and be stabilized and improved.

## 1. INTRODUCTION

Over the past few years, the areas of applying CNN has expanded in the field of computer vision. For better performance in computer vision tasks such as classification [2, 3], segmentation [4, 5] and detection [6, 7], the size of model has been developed larger and deeper. However, this trend causes that it is difficult to use CNN model in low performance systems because the huge amount of memory and computational complexity is needed.

There are various studies to solve this problem such as lightweight [8, 9], pruning [10, 11] and quantization [12, 13] etc. Knowledge distillation (KD) [14, 15] is another popular approach in model compressing. The core idea of KD is transfer large teacher model's knowledge to small student model. So, student can be trained to mimic teacher's output such as logits [14, 17] using Kullback-Leibler divergence or feature [15, 16] using Euclidean distance. So far, KD has been mainly studied to improve the performance of shallow depth student model with conventional CNN using N×N shape filter.

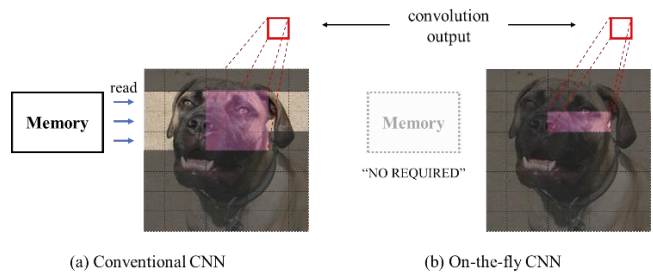

Fig. 1. An overview of data reading and convolution process difference of conventional CNN and on-the-fly CNN when N=3. (a) Conventional CNN uses N×N filter so, it needs to read image data N line pixels from memory. However, (b) on-the-fly CNN need only 3 pixels to compute the convolution output and this on-the-fly process do not required using memory.

However, we need to use 1×N filter CNN for on-the-fly system, as shown in the Fig. 1., N×N filter conventional CNN can be used for both teacher and student. And student has shallow depth than teacher for conventional KD (CKD). However, on-the-fly CNN student has same depth as teacher and uses 1×N filter for proposed PMKD. So then, student's capacity is reduced about (N−1)/N parameters. For example, if N=3, teacher uses 3×3 square filter and on-the-fly student uses 3×3 filter with a reduction of approximately 66% parameters than teacher. Besides, a student can only consider the horizontal not the vertical receptive field.

And we find that applying CKD directly to on-the-fly student causes bad performance due to the large capacity gap and on-the-fly student's limited ability. It shows a lower or slightly better performance than a baseline. To help the student overcome these limits and reach teacher stably, we propose PMKD, intermediate teacher acting as pacemaker in marathon. With PMKD, on-the-fly student model can be trained step by step to emulate reliably the original teacher.

Our contributions: (1) we studied a novel approach that the on-the-fly CNN, using 1×3 filter, can be used for low performance system requiring on-the-fly processing; (2) unlike the CKD transfer knowledge directly, we propose a novel approach adding an intermediate teacher called PMKD; (3) we solve the training instability occurred when the CKD method was used for on-the-fly student and make significant performance improvement on CIFAR and SVHN datasets.

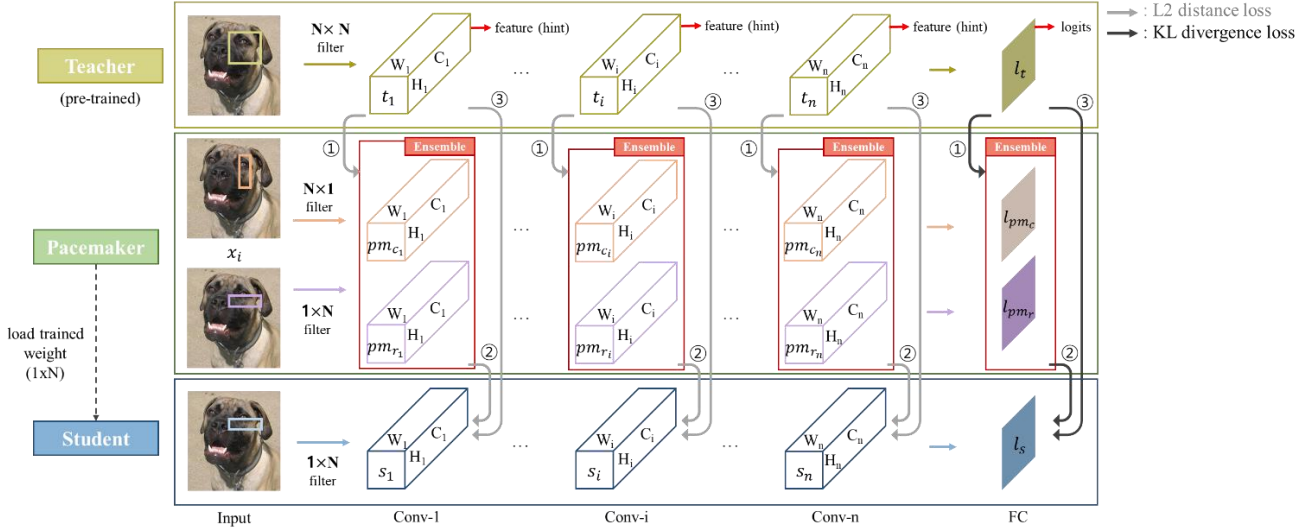

Fig. 2. The architecture of the proposed PMKD for on-the-fly CNN. PMKD is made up of the total three times distillation and all feature's W, H, C shape is totally same. The 1st distillation is done by pre-trained original teacher (OT, yellow) and the pacemaker (PM, green) and PM learns from the OT. PM is consisting of ensemble of two models using 1×N row filter and N×1 column filter, respectively. And the 2nd distillation is that PM serves as an intermediate teacher model for training student model (blue). Then, for 3rd distillation, the student model loads PM's pretrained 1×N filter model weights and learns from OT.

## 2. BACKGROUND

As briefly mentioned above, KD can be divided into two types, using logits and feature for knowledge. Hinton et al [14] propose to transfer soften class probabilities which is softmax divided with temperature parameter $\tau$ using KL-divergence. To explain the equation, let $l_t$, $l_s$ refer teacher, student logits and $P_t^\tau = softmax(l_t/\tau)$, $P_s^\tau = softmax(l_s/\tau)$ refer teacher, student soften output:

$$KL(P_t^\tau, P_s^\tau) \quad (1)$$

On the other hand, Romero et al [15] propose to transfer feature information using Euclidean distance. The feature dimension size is different from teacher and student because they use a shallower but wider teacher than student. So, they add a regressor to match the dimension size. Let the regressor $r$ and $t_i$, $s_i$ refer teacher and student feature:

$$\frac{1}{2}\|t_i - r(s_i)\|^2 \quad (2)$$

With these two distillation losses, student is trained by adding distillation losses to cross-entropy loss, let $y$ is class label:

$$H(y, P_s) \quad (3)$$

## 3. PROPOSED METHOD

One important point is that the information distortion is easily found in CKD, because it usually uses shallow depth student model than teacher. However, our on-the-fly student has the same depth as teacher. So, on-the-fly student is relieved from this defect as it does not need to use transformation method like [15] regressor or other method occurring information distortion. Additionally, on-the-fly student can learn from all hidden layer output of teacher as feature knowledge.

### 3.1. Training Loss

Proposed PMKD use both feature and logits from teacher with cross-entropy loss, that is PMKD is mixed version of [14] and modified [15] for on-the-fly CNN with adding pacemaker. As a result, the cost function of PMKD can be written as:

$$L_{FKD} = \frac{1}{n}\sum_i^n \|t_i - r_i\|^2 \quad (4)$$

$$L_{LKD} = KL(P_t^\tau, P_s^\tau) \quad (5)$$

$$L_{CE} = H(y, P_s) \quad (6)$$

$$\therefore loss = \rho L_{FKD} + \alpha L_{LKD} + (1-\alpha)L_{CE} \quad (7)$$

Above hyperparameter $\rho$ and $\alpha$ are used for balancing losses. We set $\tau = 4$, $\alpha = 0.9$ which are commonly selected value in [14, 16]. And $\rho$ is set various values over a wide range from 0.01 to 5 depending on the dataset and network in our experiments.

### 3.2. Learning Procedures

This section is to describe the detail of three procedures. It is somewhat different each phase of three phase of distillation.

**1st distillation: Teacher & Pacemaker.** This is pre-process for student training. It is conducted by teacher (N×N) and pacemaker (1×N & N×1). Teacher uses pre-trained weights and pacemaker uses randomly initialized weights. By ensemble of 1×N row filter model and N×1 column filter model, pacemaker is trained to mimic teacher.

**2nd distillation: Pacemaker & Student.** At second phase, pacemaker acts as an intermediate teacher using pre-trained weights in first phase. And a student model using 1×N filter with randomly initialized weights learns knowledge from ensemble of 1×N & N×1 dual model, pacemaker.

**3rd distillation: Student & Teacher.** The final phase is same process as CKD. An on-the-fly 1×N student learns from the original teacher using pre-trained weights in second phase.

## 4. EXPERIMENTS

### 4.1. Datasets

We conduct the experiment with classification benchmark dataset, CIFAR and SVHN. CIFAR10 and CIFAR100 [18] contain 32×32 size RGB nature image. And both have 50,000, 10,000 images for train, test set and each has 10, 100 classes. Standard data augmentation [19] is applied for CIFAR. Google's The Street View House Numbers (SVHN) [20] contains 32×32 size center closed-up RGB house digit image. It has 73,257 and 26,032 images for train and test set with 10 classes. And no data augmentation [21] is applied.

### 4.2. Networks

We conduct experiments with various network architectures: VGG [3], ResNet [22], WideResNet [23]. Note that we replace all intermediate pooling to stride process for on-the-fly on VGG. Because it is more suitable and efficient for on-the-fly CNN computational process. And we do not need to modify the other networks because these networks do not require intermediate pooling process. In addition, we avoid replacing pooling at first and last layer like [13] strategy for all networks. For VGG network, we select the first and intermediate layer features just before the channel increased and logits as knowledge. And the first and all block features and logits for the series of residual blocks models.

### 4.3. Settings

For all experiment, we run 200 epochs and set the batch size 128 using SGD with momentum 0.9 and weight decay 1e-5. The initial learning rate is 0.1 and dropping 0.2 per [60, 120, 160] epochs. And we set the filter size of N=3 for all model teacher, pacemaker and student. So, teacher used only 3×3 filter, pacemaker use 1×3 and 3×1 filter for ensemble and student use only 1×3 filter for intermediate layers. To sum up, three kinds of filter are used. To match the same spatial size, teacher use zero padding 1 while student use (0, 1) and both have stride 1. And we repeated the same experiment 5 times and report the average value of the results.

### 4.4. Result Analysis

*4.4.1. Problem Statement*

As below Table 1, 2, 3, applying CKD to on-the-fly CNN occurs bad performance or slight improvement on CIFAR and SVHN datasets. It's because of large capacity gap between teacher and student [24]. And an on-the-fly CNN student model, can only consider horizontal field. Due to these weak points, it is very hard to student to learn the knowledge directly from teacher. So, the bad performance and instable train problems can be occurred, and it is hard to solve these problems on student itself.

*4.4.2. Performance improvement by proposed PMKD*

We can find the problem on below Table 1, 2, 3. For example, as shown in Table 1, VGG16 model on CIFAR10 shows underperformance when directly applied CKD such as KD [14] and FitNet [15] than baseline −2.73% and −1.69%, respectively. And note that if we use both KD and FitNet together, it shows worst performance −5.44%. This result means that due to low capacity, on-the-fly student cannot acquire much information at once. However, when we add proposed PM, it shows improved performance +0.88% and +0.73%, respectively. Note that PMKD, using PM with KD and FitNet, shows best performance improved +1.46%. And we can find similar results on Table 2, 3 too.

Table 1: Experiments results of top 1 accuracy on CIFAR10. For easy understanding, we denote ↓ when student underperform than baseline, not applied CKD, ↑ as improved performance, and bold as best performance. These denotations are used equally on Table 2, 3.

| Model | VGG13 (BN) | VGG16 (BN) | VGG19 (BN) | ResNet18 | ResNet34 | ResNet50 | WRN 10-10 | WRN 16-8 | WRN 28-6 | WRN 40-4 |
|---|---|---|---|---|---|---|---|---|---|---|
| Teacher | 92.93 | 93.63 | 93.60 | 94.64 | 94.23 | 94.34 | 93.68 | 94.67 | 95.34 | 95.45 |
| Baseline | 69.72 | 70.96 | 70.20 | 82.15 | 82.24 | 81.62 | 71.82 | 73.66 | 74.20 | 74.22 |
| KD | 68.35 (↓) | 68.23 (↓) | 69.08 (↓) | 82.63 (↑) | 82.63 (↑) | 82.58 (↑) | 71.14 (↓) | 73.99 (↑) | 73.97 (↓) | 73.82 (↓) |
| FitNet | 70.84 (↑) | 69.27 (↓) | 69.39 (↓) | 82.09 (↓) | 81.59 (↓) | 82.09 (↑) | 70.89 (↓) | 72.88 (↓) | 74.03 (↓) | 74.49 (↑) |
| KD + FitNet | 65.31 (↓) | 65.52 (↓) | 64.89 (↓) | 81.57 (↓) | 80.69 (↓) | 79.07 (↓) | 68.05 (↓) | 71.83 (↓) | 72.69 (↓) | 72.31 (↓) |
| PM + KD | 71.27 (↑) | 71.84 (↑) | 71.02 (↑) | 82.90 (↑) | 83.20 (↑) | 81.94 (↑) | 72.11 (↑) | 74.93 (↑) | 75.78 (↑) | 75.51 (↑) |
| PM + FitNet | 70.96 (↑) | 71.69 (↑) | 70.45 (↑) | 82.74 (↑) | 82.66 (↑) | 81.79 (↑) | 72.53 (↑) | 75.01 (↑) | 75.54 (↑) | 75.34 (↑) |
| PMKD (ours) | **71.44** (↑) | **72.42** (↑) | **72.19** (↑) | **83.44** (↑) | **83.67** (↑) | **82.82** (↑) | **73.26** (↑) | **75.85** (↑) | **76.03** (↑) | **75.91** (↑) |

Table 2: Experiments results of top 1 accuracy on CIFAR100.

| Model | VGG13 (BN) | VGG16 (BN) | VGG19 (BN) | ResNet18 | ResNet34 | ResNet50 | WRN 10-10 | WRN 16-8 | WRN 28-6 | WRN 40-4 |
|---|---|---|---|---|---|---|---|---|---|---|
| Teacher | 72.13 | 72.66 | 71.18 | 74.52 | 76.44 | 75.67 | 74.42 | 77.43 | 78.68 | 78.45 |
| Baseline | 41.27 | 43.52 | 40.76 | 57.13 | 57.40 | 53.55 | 47.46 | 48.70 | 48.78 | 49.19 |
| KD | 40.82 (↓) | 42.34 (↓) | 38.94 (↓) | 57.22 (↑) | 57.96 (↑) | 52.63 (↓) | 45.51 (↓) | 48.03 (↓) | 47.46 (↓) | 47.92 (↓) |
| FitNet | 41.51 (↑) | 41.86 (↓) | 38.13 (↓) | 56.92 (↓) | 57.55 (↑) | 51.88 (↓) | 46.75 (↓) | 47.88 (↓) | 48.33 (↓) | 48.71 (↓) |
| KD + FitNet | 39.73 (↓) | 41.26 (↓) | 36.42 (↓) | 56.48 (↓) | 57.32 (↓) | 50.04 (↓) | 46.49 (↓) | 48.00 (↓) | 49.28 (↑) | 48.50 (↓) |
| PM + KD | 45.17 (↑) | 46.61 (↑) | 43.61 (↑) | 59.87 (↑) | 59.52 (↑) | 55.07 (↑) | 50.77 (↑) | 51.06 (↑) | 51.12 (↑) | 51.39 (↑) |
| PM + FitNet | 44.80 (↑) | 46.66 (↑) | 42.89 (↑) | 60.46 (↑) | 58.73 (↑) | 54.48 (↑) | 51.24 (↑) | 50.81 (↑) | 50.82 (↑) | 51.21 (↑) |
| PMKD (ours) | **45.93** (↑) | **47.39** (↑) | **44.24** (↑) | **61.11** (↑) | **60.49** (↑) | **56.95** (↑) | **51.78** (↑) | **52.54** (↑) | **52.44** (↑) | **53.89** (↑) |

Table 3: Experiments results of top 1 accuracy on SVHN.

| Model | VGG13 (BN) | VGG16 (BN) | VGG19 (BN) | ResNet18 | ResNet34 | ResNet50 | WRN 10-10 | WRN 16-8 | WRN 28-6 | WRN 40-4 |
|---|---|---|---|---|---|---|---|---|---|---|
| Teacher | 95.08 | 95.36 | 95.53 | 95.67 | 96.20 | 96.09 | 96.78 | 97.21 | 97.23 | 97.00 |
| Baseline | 69.48 | 69.78 | 68.97 | 77.35 | 77.22 | 77.10 | 51.30 | 52.33 | 52.67 | 53.92 |
| KD | 70.04 (↑) | 70.40 (↑) | 69.58 (↑) | 74.98 (↓) | 74.67 (↓) | 73.43 (↓) | 48.59 (↓) | 50.04 (↓) | 50.38 (↓) | 50.69 (↓) |
| FitNet | 68.71 (↓) | 69.16 (↓) | 68.09 (↓) | 74.59 (↓) | 73.21 (↓) | 75.07 (↓) | 47.27 (↓) | 51.52 (↓) | 51.93 (↓) | 52.68 (↓) |
| KD + FitNet | 69.83 (↑) | 69.22 (↓) | 65.29 (↓) | 68.83 (↓) | 70.03 (↓) | 76.19 (↓) | 45.76 (↓) | 51.45 (↓) | 51.25 (↓) | 51.87 (↓) |
| PM + KD | 72.14 (↑) | 72.17 (↑) | 77.63 (↑) | 79.56 (↑) | 79.97 (↑) | 79.25 (↑) | 51.47 (↑) | 52.31 (↑) | 53.36 (↑) | 54.32 (↑) |
| PM + FitNet | 70.92 (↑) | 72.04 (↑) | 78.35 (↑) | 78.42 (↑) | 79.24 (↑) | 78.75 (↑) | 51.86 (↑) | 52.93 (↑) | 53.94 (↑) | 54.75 (↑) |
| PMKD (ours) | **71.64** (↑) | **72.76** (↑) | **78.67** (↑) | **80.67** (↑) | **81.04** (↑) | **80.33** (↑) | **52.23** (↑) | **53.19** (↑) | **54.31** (↑) | **55.42** (↑) |

### 4.4.3. Training stability

Applying CKD directly to on-the-fly student occurs very unstable results that the variation between each result was very large. When applying KD [14] to WRN 28-6 model on CIFAR100 shows wide range performance [44.69, 46.53, 47.84, 48.73, 49.51] for 5 experiments. So, the average is 47.46, shown in Table 2, and the deviation is [−2.77, −0.93, +0.38, +1.27, +2.05]. However, by applying PM, it shows result [50.72, 50.95, 51.02, 51.23, 51.63], the average is 51.12 and the deviation is [−0.4, −0.17, −0.1, +0.11, +0.51]. As a result, the deviation range is reduced by the applying PM. So, we can say that our proposed method PMKD makes stable training environment to on-the-fly CNN student.

### 4.4.4. Why should the pacemaker be an ensemble?

Intuitively, using only 1×N model as PM will not help for on-the-fly student, because it is the same as student. As shown in Table 4, when PM is consisting of ensemble models, the on-the-fly student shows better performance than only use 1×N model. So, we can make sure that PM should be ensemble of 1×N & N×1 dual model than only uses N×1 filter model.

Table 4: PMKD result of *N×1* vs *1×N & N×1* for PM, when N=3.

| Dataset | Model / Pacemaker | VGG13 (BN) | ResNet18 | WRN 10-10 |
|---|---|---|---|---|
| CIFAR10 | 3 × 1 | 70.97 | 82.80 | 72.24 |
| | 1 × 3 & 3 × 1 | **71.44** | **83.44** | **73.26** |
| CIFAR100 | 3 × 1 | 42.23 | 59.56 | 49.39 |
| | 1 × 3 & 3 × 1 | **45.93** | **61.11** | **51.78** |
| SVHN | 3 × 1 | 70.86 | 78.72 | 51.83 |
| | 1 × 3 & 3 × 1 | **71.64** | **80.67** | **52.23** |

## 5. CONCLUSION

In this paper, we study to use the on-the-fly CNN as student model which use 1×N filter for low performance systems [1] and focus on improving the student model performance using knowledge distillation. We find that the student can show underperformance or slightly better than baseline, when the conventional knowledge distillation is applied directly to on-the-fly CNN. This problem is due to large capacity gap [24] between N×N teacher and on-the-fly student which has the limit that it can only consider the horizontal receptive field. So, to solve the problem and make the on-the-fly student improved, we present pacemaker knowledge distillation, intermediate temporary teacher. And we find out that on-the-fly student shows better performance when pacemaker is ensemble of 1×N & N×1 dual model rather than only N×1 model. We believe that by applying proposed method pacemaker to knowledge distillation, low capacity on-the-fly student model shows significantly improvement that directly use knowledge distillation to on-the-fly student model.

## 6. ACKNOWLEDGEMENT

This project is partially supported by Samsung Electronics CO., LTD and partially supported by the Technology Innovation Program (20000316, Scene Understanding and Threat Assessment based on Deep Learning for Automatic Emergency Steering) funded By the MOTIE, Korea and KIAT grant funded by the Korea Government(MOTIE) (N0002428, The Competency Development Program for Industry Specialist)